\documentclass[runningheads]{llncs}
\usepackage{graphicx}
\usepackage{comment}
\usepackage{amsmath,amssymb} 
\usepackage{color}
\usepackage{adjustbox}
\usepackage{subfig}
\usepackage{algorithm}
\usepackage{algorithmic}
\usepackage{multirow}
\usepackage{enumerate}
\usepackage{verbatim}
\usepackage[symbol]{footmisc}

\usepackage{hyperref}
\newcommand{\etal}{\textit{et al}. }
\newcommand{\ie}{\textit{i}.\textit{e}., }
\newcommand{\eg}{\textit{e}.\textit{g}. }

\newlength\savewidth\newcommand\shline{\noalign{\global\savewidth\arrayrulewidth
		\global\arrayrulewidth 1pt}\hline\noalign{\global\arrayrulewidth\savewidth}}
\newcommand*\samethanks[1][\value{footnote}]{\footnotemark[#1]}

\definecolor{gray}{rgb}{0.5,0.5,0.5}

\begin{document}
\pagestyle{headings}
\mainmatter
\def\ECCVSubNumber{2085}  

\title{Deep Image Clustering with Category-Style Representation} 

\titlerunning{Deep Image Clustering with Category-Style Representation}
%
\author{Junjie Zhao\inst{1}\thanks{Equal contribution and the work was done at Tencent Jarvis Lab.} \and
Donghuan Lu\inst{2}\samethanks[1] \and
Kai Ma\inst{2} \and
Yu Zhang\inst{1}\thanks{Y. Zhang and Y. Zheng are the corresponding authors.} \and 
Yefeng Zheng\inst{2}\samethanks[2]}

\authorrunning{J. Zhao, D. Lu, K. Ma, Y. Zhang and Y. Zheng}
%
\institute{School of Computer Science and Engineering, Southeast University, Nanjing, China \\
\email{\{kamij.zjj,zhang\_yu\}@seu.edu.cn} \and
Tencent Jarvis Lab, Shenzhen, China \\
\email{\{caleblu,kylekma,yefengzheng\}@tencent.com}}
\maketitle

\begin{abstract}
Deep clustering which adopts deep neural networks to obtain optimal representations for clustering has been widely studied recently. In this paper, we propose a novel deep image clustering framework to learn a category-style latent representation in which the category information is disentangled from image style and can be directly used as the cluster assignment. To achieve this goal, mutual information maximization is applied to embed relevant information in the latent representation. Moreover, augmentation-invariant loss is employed to disentangle the representation into category part and style part. Last but not least, a prior distribution is imposed on the latent representation to ensure the elements of the category vector can be used as the probabilities over clusters. Comprehensive experiments demonstrate that the proposed approach outperforms state-of-the-art methods significantly on five public datasets.\footnote{Project address: \href{https://github.com/sKamiJ/DCCS}{\textcolor{magenta}{https://github.com/sKamiJ/DCCS}}.}
\end{abstract}

\section{Introduction}
\label{sec:intro}
Clustering is a widely used technique in many fields, such as machine learning, data mining and statistical analysis. It aims to group objects `similar' to each other into the same set and `dissimilar' ones into different sets. Unlike supervised learning methods, clustering approaches should be oblivious to ground truth labels. Conventional methods, such as K-means~\cite{macqueen1967some} and spectral clustering~\cite{ng2002spectral}, require feature extraction to convert data to a more discriminative form. Domain knowledge could be useful to determine more appropriate feature extraction strategies in some cases. But for many high-dimensional problems (\eg images), manually designed feature extraction methods can easily lead to inferior performance.

Because of the powerful capability of deep neural networks to learn non-linear mapping, a lot of deep learning based clustering methods have been proposed recently. Many studies attempt to combine deep neural networks with various kinds of clustering losses~\cite{xie2016unsupervised,guo2017deep,ghasedi2017deep} to learn more discriminative yet low-dimensional latent representations. To avoid trivially learning some arbitrary representations, most of those methods also minimize a reconstruction~\cite{guo2017deep} or generative~\cite{mukherjee2019clustergan} loss as an additional regularization. However, there is no substantial connection between the discriminative ability and the generative ability of the latent representation. The aforementioned regularization turns out to be less relevant to clustering and forces the latent representation to contain unnecessary generative information, which makes the network hard to train and could also affect the clustering performance.

In this paper, instead of using a decoder/generator to minimize the reconstruction/generative loss, we use a discriminator to maximize the mutual information~\cite{hjelm2018learning} between input images and their latent representations in order to retain discriminative information. To further reduce the effect of irrelevant information, the latent representation is divided into two parts, \ie the category~(or cluster) part and the style part, where the former one contains the distinct identities of images~(inter-class difference) while the latter one represents style information~(intra-class difference). Specifically, we propose to use data augmentation to disentangle the category representation from style information, based on the observation~\cite{wu2019deep,ji2019invariant} that appropriate augmentation should not change the image category.

Moreover, many deep clustering methods require additional operations~\cite{xie2016unsupervised,shaham2018spectralnet} to group the latent representation into different categories. But their distance metrics are usually predefined and may not be optimal. In this paper, we impose a prior distribution~\cite{mukherjee2019clustergan} on the latent representation to make the category part closer to the form of a one-hot vector, which can be directly used to represent the probability distribution of the clusters.

In summary, we propose a novel approach, \textbf{D}eep \textbf{C}lustering with \textbf{C}ategory-\textbf{S}tyle representation~(DCCS) for unsupervised image clustering. The main contributions of this study are four folds:
\begin{itemize}
    \item We propose a novel end-to-end deep clustering framework to learn a latent category-style representation whose values can be used directly for the cluster assignment.
    \item We show that maximizing the mutual information is enough to prevent the network from learning arbitrary representations in clustering.
    \item We propose to use data augmentation to disentangle the category representation~(inter-class difference) from style information~(intra-class difference).
    \item Comprehensive experiments demonstrate that the proposed DCCS approach outperforms state-of-the-art methods on five commonly used datasets, and the effectiveness of each part of the proposed method is evaluated and discussed in thorough ablation studies.
\end{itemize}

\section{Related Work}
In recent years, many deep learning based clustering methods have been proposed. Most approaches~\cite{guo2017deep,ghasedi2017deep,yang2019deep} combined autoencoder~\cite{bengio2007greedy} with traditional clustering methods by minimizing reconstruction loss as well as clustering loss. For example, Jiang \etal\cite{jiang2016variational} combined a variational autoencoder~(VAE)~\cite{kingma2013auto} for representation learning with a Gaussian mixture model for clustering. Yang \etal\cite{yang2019dgg} also adopted the Gaussian mixture model as the prior in VAE, and incorporated a stochastic graph embedding to handle data with complex spread. Although the usage of the reconstruction loss can embed the sample information into the latent space, the encoded latent representation may not be optimal for clustering.  

Other than autoencoder, Generative Adversarial Network~(GAN)~\cite{goodfellow2014generative} has also been employed for clustering~\cite{chen2016infogan}. In ClusterGAN~\cite{mukherjee2019clustergan}, Mukherjee \etal also imposed a prior distribution on the latent representation, which was a mixture of one-hot encoded variables and continuous latent variables. Although their representations share a similar form to ours, their one-hot variables cannot be used as the cluster assignment directly due to the lack of proper disentanglement. Additionally, ClusterGAN consisted of a generator~(or a decoder) to map the random variables from latent space to image space, a discriminator to ensure the generated samples close to real images and an encoder to map the images back to the latent space to match the random variables. Such a GAN model is known to be hard to train and brings irrelevant generative information to the latent representations. To reduce the complexity of the network and avoid unnecessary generative information, we directly train an encoder by matching the aggregated posterior distribution of the latent representation to the prior distribution.

To avoid the usage of additional clustering, some methods directly encoded images into latent representations whose elements can be treated as the probabilities over clusters. For example, Xu \etal\cite{ji2019invariant} maximized pair-wise mutual information of the latent representations extracted from an image and its augmented version. This method achieved good performance on both image clustering and segmentation, but its batch size must be large enough~(more than 700 in their experiments) so that samples from different clusters were almost equally distributed in each batch. Wu \etal\cite{wu2019deep} proposed to learn one-hot representations by exploring various correlations of the unlabeled data, such as local robustness and triple mutual information. However, their computation of mutual information required pseudo-graph to determine whether images belonged to the same category, which may not be accurate due to the unsupervised nature of clustering. To avoid this issue, we maximized the mutual information between images and their own representations instead of representations encoded from images with the same predicted category. 
 
\section{Method}
\begin{figure}[t]
    \centering
    \includegraphics[width=0.95\linewidth]{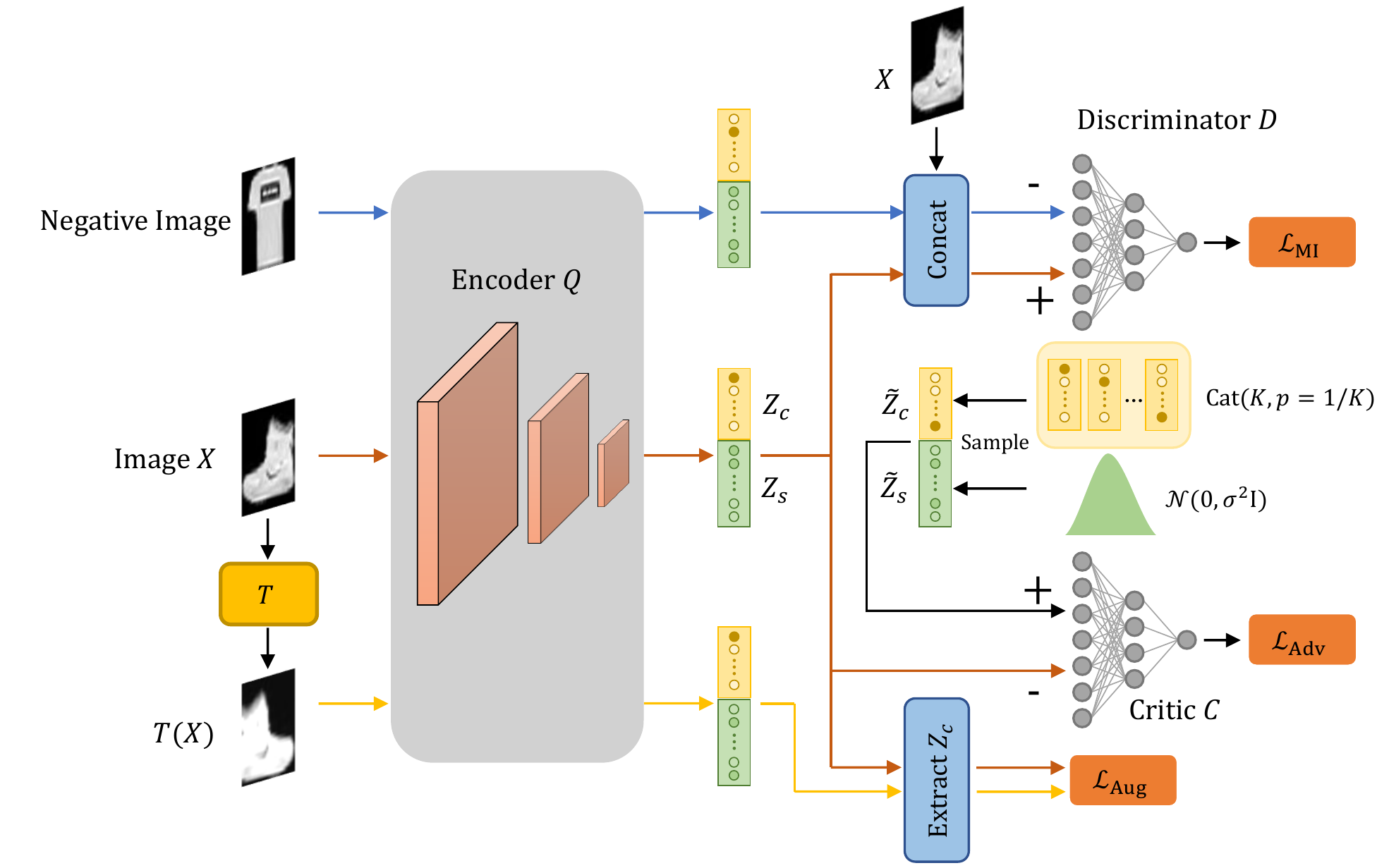}
    \caption{The overall framework of the proposed DCCS method. The encoder $Q$ converts the image $X$ into a latent representation $Z=(Z_c,Z_s)$. The discriminator $D$ maximizes the mutual information between $X$ and $Z$, while the critic $C$ imposes a prior distribution on $Z$ to make $Z_c$ closer to the form of a one-hot vector and constrain the range of $Z_s$. $Z_c$ is also regularized to be invariant to data augmentation $T$}
    \label{fig:framework}
\end{figure}

As stated in the introduction, the proposed DCCS approach aims to find an appropriate encoder $Q$ to convert the input image $X$ into a latent representation $Z$, which consists of disentangled category and style information. To be more precise, the encoded latent representation $Z$ consists of a softmax-activated vector $Z_c$ and a linear-activated vector $Z_s$, \ie $Z=(Z_c,Z_s)$, where $Z_c$ represents the probabilities of $X$ belonging to each class and $Z_s$ represents the intra-class style information. To achieve this, three regularization strategies are applied to constrain the latent representation as detailed in the following three sections, and the framework is shown in Fig.~\ref{fig:framework}. To clarify notations, we use upper case letters~(\eg $X$) for random variables, lower case letters~(\eg $x$) for their values and calligraphic letters~(\eg $\mathcal{X}$) for sets. The probability distributions are denoted with upper case letters~(\eg $P(X)$), and the corresponding densities are denoted with lower case letters~(\eg $p(x)$).

\subsection{Maximize Mutual Information}
\label{sec:maximize_mutual_information}
Because of the powerful capability to fit training data with complex non-linear transformations, the encoder of deep neural networks can easily map input images to arbitrary representations if without proper constraints thus losing relevant information for proceeding the target clustering task. To retain the essential information of each image and learn better discriminative latent representations, a discriminator $D$ is introduced to maximize the mutual information $I(X,Z)$ between the input image $X$ and its encoded latent representation $Z$. Based on information theory, $I(X,Z)$ takes the following form:
\begin{align}
	I(X,Z)&=\iint q(z|x)p_X(x)\log\frac{q(z|x)p_X(x)}{q_Z(z)p_X(x)} dxdz \\
	&=\text{KL}(Q(Z|X)P_X(X)\|Q_Z(Z)P_X(X))
\end{align}
where $Q(Z|X)$ is the encoding distribution, $P_X$ is the prior distribution of the images, $Q_Z=\mathbb{E}_{P_X}[Q(Z|X)]$ is the aggregated posterior distribution of the latent representation and $\text{KL}(\cdot\|\cdot)$ is the KL-divergence. In this original formulation, however, KL-divergence is unbounded and maximizing it may lead to an unstable training result. Following~\cite{hjelm2018learning}, we replace KL-divergence with JS-divergence to estimate the mutual information:
\begin{equation}
	I^{(\text{JSD})}(X,Z)=\text{JS}(Q(Z|X)P_X(X),Q_Z(Z)P_X(X)).
	\label{eq:JS-divergence}
\end{equation}
According to~\cite{nowozin2016f,goodfellow2014generative}, JS-divergence between two arbitrary distributions $P(X)$ and $Q(X)$ can be estimated by a discriminator $D$:
\begin{equation}
\begin{aligned}
	\text{JS}(P(X),Q(X))=\frac{1}{2}\max_{D}\{&\mathbb{E}_{X\sim P(X)}[\log S(D(X))] \\
		 + &\mathbb{E}_{X\sim Q(X)}[\log(1-S(D(X)))]\} + \log 2
\end{aligned}
\end{equation}
where $S$ is the sigmoid function. Replacing $P(X)$ and $Q(X)$ with $Q(Z|X)P_X(X)$ and $Q_Z(Z)P_X(X)$, the mutual information can be maximized by:
\begin{equation}
\begin{aligned}
	\frac{1}{2}\max_{Q,D}\{&\mathbb{E}_{(X,Z)\sim Q(Z|X)P_X(X)}[\log S(D(X,Z))] \\
	+ &\mathbb{E}_{(X,Z)\sim Q_Z(Z)P_X(X)}[\log(1-S(D(X,Z)))]\} + \log 2.
\end{aligned}
\end{equation}
Accordingly, the mutual information loss function can be defined as:
\begin{equation}
\begin{aligned}
	\mathcal{L}_{\text{MI}}= - (&\mathbb{E}_{(X,Z)\sim Q(Z|X)P_X(X)}[\log S(D(X,Z))] \\
	+ &\mathbb{E}_{(X,Z)\sim Q_Z(Z)P_X(X)}[\log(1-S(D(X,Z)))])
\end{aligned}
\end{equation}
where $Q$ and $D$ are jointly optimized.

With the concatenation of $X$ and $Z$ as input, minimizing $\mathcal{L}_{\text{MI}}$ can be interpreted as to determine whether $X$ and $Z$ are correlated. For discriminator $D$, an image $X$ along with its representation is a positive sample while $X$ along with a representation encoded from another image is a negative sample. As aforementioned, many deep clustering methods use the reconstruction loss or generative loss to avoid arbitrary representations. However, it allows the encoded representation to contain unnecessary generative information and makes the network, especially GAN, hard to train. The mutual information maximization only instills necessary discriminative information into the latent space and experiments in Section~\ref{sec:experiment} confirm that it leads to better performance.

\subsection{Disentangle Category-Style Information}
\label{sec:disentangle}
As previously stated, we expect the latent category representation $Z_c$ only contains the categorical cluster information while all the style information is represented by $Z_s$. To achieve such a disentanglement, an augmentation-invariant regularization term is introduced based on the observation that certain augmentation should not change the category of images.

Specifically, given an augmentation function $T$ which usually includes geometric transformations~(\eg scaling and flipping) and photometric transformations~(\eg changing brightness or contrast), $Z_c$ and $Z'_c$ encoded from $X$ and $T(X)$ should be identical while all the appearance differences should be represented by the style variables. Because the elements of $Z_c$ represent the probabilities over clusters, the KL-divergence is adopted to measure the difference between $Q(Z_c|X)$ and $Q(Z_c|T(X))$. The augmentation-invariant loss function for the encoder $Q$ can be defined as:
\begin{equation}
	\mathcal{L}_{\text{Aug}}=\text{KL}(Q(Z_c|X)\|Q(Z_c|T(X))).
\end{equation}

\subsection{Match to Prior Distribution}
\label{sec:match_to_prior}
There are two potential issues with the aforementioned regularization terms: the first one is that the category representation cannot be directly used as the cluster assignment, therefore additional operations are still required to determine the clustering categories; the second one is that the augmentation-invariant loss may lead to a degenerate solution, \ie assigning all images into a few clusters, or even the same cluster. In order to resolve these issues, a prior distribution $P_Z$ is imposed on the latent representation $Z$. 

Following~\cite{mukherjee2019clustergan}, a categorical distribution $\tilde{Z}_c\sim \text{Cat}(K,p=1/K)$ is imposed on $Z_c$, where $\tilde{Z}_c$ is a one-hot vector and $K$ is the number of categories that the images should be clustered into. A Gaussian distribution $\tilde{Z}_s\sim \mathcal{N}(0, \sigma ^2\mathbf{I})$~(typically $\sigma =0.1$) is imposed on $Z_s$ to constrain the range of style variables.

As aforementioned, ClusterGAN~\cite{mukherjee2019clustergan} uses the prior distribution to generate random variables, applies a GAN framework to train a proper decoder and then learns an encoder to match the decoder. To reduce the complexity of the network and avoid unnecessary generative information, we directly train the encoder by matching the aggregated posterior distribution $Q_Z=\mathbb{E}_{P_X}[Q(Z|X)]$ to the prior distribution $P_Z$. Experiments demonstrate that such a strategy can lead to better clustering performance.

To impose the prior distribution $P_Z$ on $Z$, we minimize the Wasserstein distance~\cite{arjovsky2017wasserstein} $W(Q_Z,P_Z)$ between $Q_Z$ and $P_Z$, which can be estimated by:
\begin{equation}
	\max_{C\in \mathcal{C}} \{\mathbb{E}_{\tilde{Z} \sim P_Z}[C(\tilde{Z})] - \mathbb{E}_{Z\sim Q_Z}[C(Z)]\}
	\label{eq:wasserstein_distance}
\end{equation}
where $\mathcal{C}$ is the set of 1-Lipschitz functions. Under the optimal critic $C$~(denoted as \textit{discriminator} in vanilla GAN), minimizing Eq.~\ref{eq:wasserstein_distance} with respect to the encoder parameters also minimizes $W(Q_Z,P_Z)$: 
\begin{equation}
	\min_{Q}\max_{C\in \mathcal{C}} \{\mathbb{E}_{\tilde{Z} \sim P_Z}[C(\tilde{Z})] - \mathbb{E}_{Z\sim Q_Z}[C(Z)]\}.
\end{equation}
For optimization, the gradient penalty~\cite{gulrajani2017improved} is introduced to enforce the Lipschitz constraint on the critic. The adversarial loss functions for the encoder $Q$ and the critic $C$ can be defined as:
\begin{equation}
	\mathcal{L}_{\text{Adv}}^Q= - \mathbb{E}_{Z\sim Q_Z}[C(Z)]
\end{equation}
\begin{equation}
	\mathcal{L}_{\text{Adv}}^C= \mathbb{E}_{Z\sim Q_Z}[C(Z)] - \mathbb{E}_{\tilde{Z} \sim P_Z}[C(\tilde{Z})] + \lambda \mathbb{E}_{\hat{Z} \sim P_{\hat{Z}}}[(\|\nabla_{\hat{Z}}C(\hat{Z})\|_2-1)^2]
\end{equation}
where $\lambda$ is the gradient penalty coefficient, $\hat{Z}$ is the latent representation sampled uniformly along the straight lines between pairs of latent representations sampled from $Q_Z$ and $P_Z$, and $(\|\nabla_{\hat{Z}}C(\hat{Z})\|_2-1)^2$ is the one-centered gradient penalty. $Q$ and $C$ are optimized alternatively.

Note that the reason why we use Wasserstein distance instead of \textit{f}-divergence is that Wasserstein distance is continuous everywhere and differentiable almost everywhere. Such a critic is able to provide meaningful gradients for the encoder even with an input containing discrete variables. On the other hand, the loss of the critic can be viewed as an estimation of $W(Q_Z,P_Z)$ to determine whether the clustering progress has converged or not, as shown in Section~\ref{sec:training_progress}.

\begin{figure}[t]
	\centering
	  \subfloat[]{\label{fig:tsne_prior}%
		\includegraphics[width=.5\linewidth]{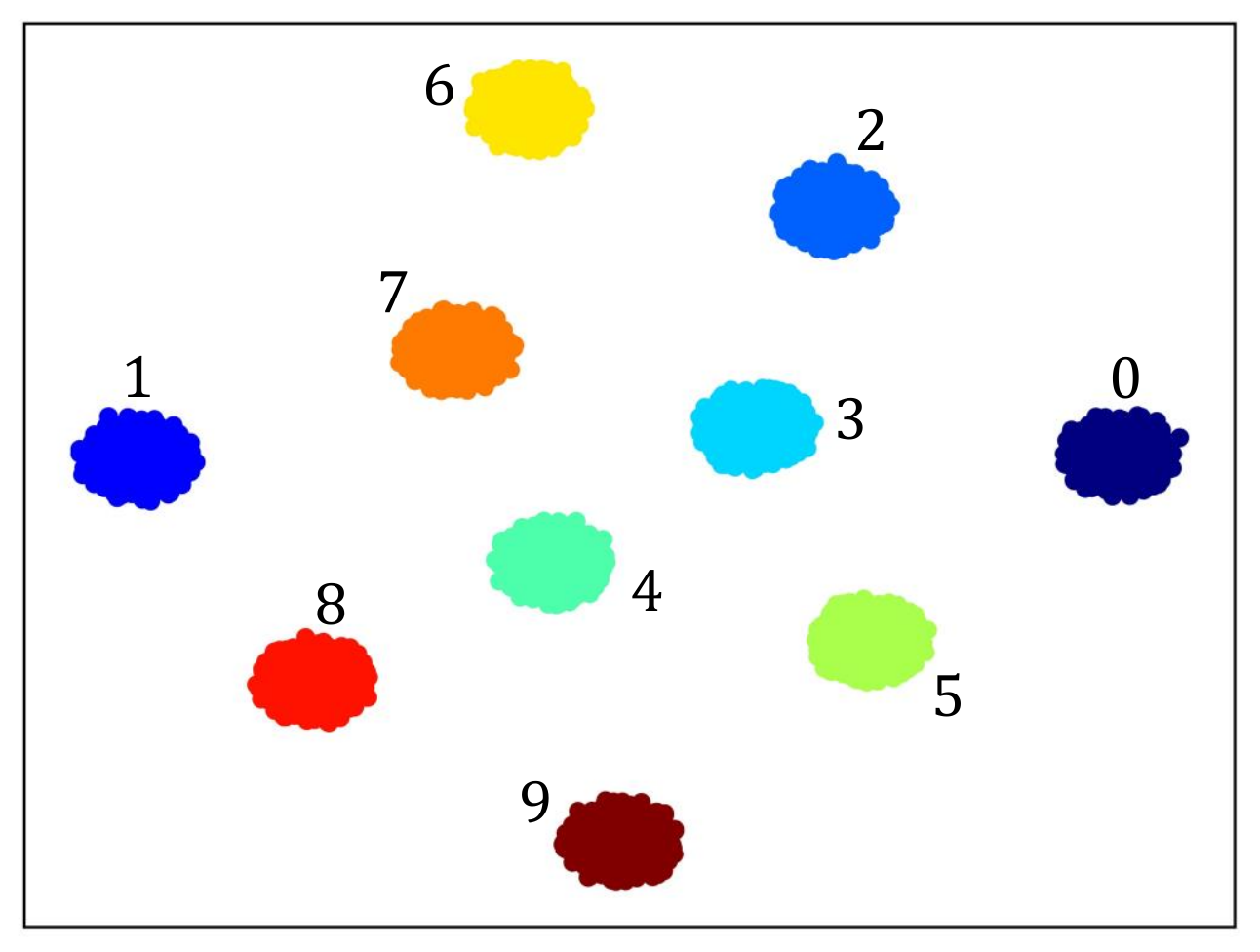}}\hfill
	  \subfloat[]{\label{fig:tsne_mnist}%
		\includegraphics[width=.5\linewidth]{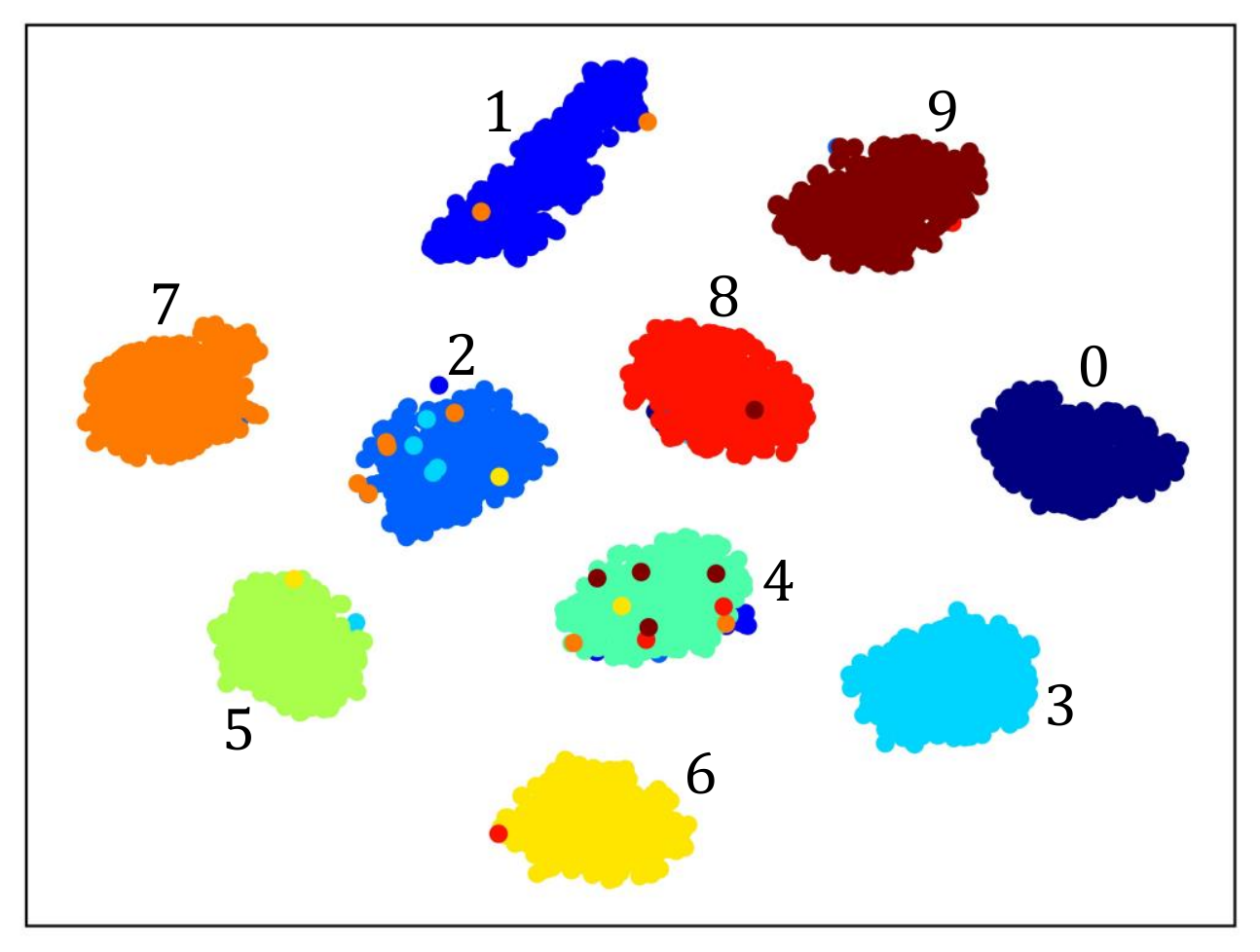}}\hfill
	  \caption{The t-SNE visualization of the latent representations of MNIST dataset. The dimensions of $Z_c$ and $Z_s$ are set as 10 and 50, respectively. Here, (a) shows the prior representation $\tilde{Z}$ sampled from $P_Z$, the numbers 0-9 represent different categories, while (b) demonstrates the encoded representation $Z$. Each point represents a latent representation and the color refers to its ground truth label}
	  \label{fig:tsne_prior_and_mnist}
\end{figure}

Fig.~\ref{fig:tsne_prior} shows the t-SNE~\cite{maaten2008visualizing} visualization of the prior representation $\tilde{Z}=(\tilde{Z}_c,\tilde{Z}_s)$ with points being colored based on $\tilde{Z}_c$. It shows that the representations sampled from the prior distribution can be well clustered based on $\tilde{Z}_c$ while $\tilde{Z}_s$ represents the intra-class difference. After imposing the prior distribution on $Z$ as displayed in Fig.~\ref{fig:tsne_mnist}, the encoded latent representations show a similar pattern as the prior representations, therefore the cluster assignment can be easily achieved by using argmax over $Z_c$.

\subsection{The Unified Model}
As shown in Fig.~\ref{fig:framework}, the network of DCCS consists of three parts: the encoder $Q$ to convert images into latent representations, the discriminator $D$ to maximize the mutual information and the critic $C$ to impose the prior distribution. The objective functions for encoder $Q$, discriminator $D$ and critic $C$ are defined as:
\begin{equation}
	\mathcal{L}^Q=\beta_{\text{MI}}\mathcal{L}_{\text{MI}}+\beta_{\text{Aug}}\mathcal{L}_{\text{Aug}}+\beta_{\text{Adv}}\mathcal{L}_{\text{Adv}}^Q
	\label{eq:loss_for_Q}
\end{equation}
\begin{equation}
	\mathcal{L}^D=\beta_{\text{MI}}\mathcal{L}_{\text{MI}}
	\label{eq:loss_for_D}
\end{equation}
\begin{equation}
	\mathcal{L}^C=\beta_{\text{Adv}}\mathcal{L}_{\text{Adv}}^C
	\label{eq:loss_for_C}
\end{equation}
where $\beta_{\text{MI}}$, $\beta_{\text{Aug}}$ and $\beta_{\text{Adv}}$ are the weights used to balance each term. 

As described in Algorithm~\ref{alg:DCCS}, the parameters of $Q$ and $D$ are jointly updated while the parameters of $C$ are trained separately. Note that because $Q$ is a deterministic encoder, \ie $Q(Z|X=x)=\delta_{\mu(x)}$, where $\delta$ denotes Dirac-delta and $\mu(x)$ is a deterministic mapping function, sampling $z^i$ from $Q(Z|X=x^i)$ is equivalent to assign $z^i$ with $\mu(x^i)$.

\renewcommand{\algorithmicrequire}{\textbf{Input:}}
\renewcommand{\algorithmicensure}{\textbf{Output:}}
\begin{algorithm}[t]
	\caption{Deep Clustering with Category-Style Representation}
	\label{alg:DCCS}
	\begin{algorithmic}[1]
		\REQUIRE  Dataset $\mathcal{X} = \{x^i\}_{i=1}^{N}$, $\theta_Q$, $\theta_D$, $\theta_C$ initial parameters of encoder $Q$, discriminator $D$ and critic $C$, the dimensions of $Z_c$ and $Z_s$, hyper-parameters $\sigma$, $\lambda$, $\beta_{\text{MI}}$, $\beta_{\text{Aug}}$, $\beta_{\text{Adv}}$, augmentation function $T$, the number of critic iterations per encoder iteration $n_{\text{critic}}$, batch size $m$.\\
		\STATE \textbf{while} $\mathcal{L}^C$ not converged \textbf{do} \\
		\STATE \quad \textbf{for} $t=1,...,n_{\text{critic}}$ \textbf{do} \\
		\STATE \qquad Sample $\{x^i\}_{i=1}^{m}$ from $\mathcal{X}$, $\{\tilde{z}^i\}_{i=1}^{m}$ from $P_Z$, $\{\epsilon^i\}_{i=1}^{m}$ from $U[0,1]$; \\
		\STATE \qquad Sample $z^i$ from $Q(Z|X=x^i)$ for $i=1,...,m$; \\
		\STATE \qquad Compute $\hat{z}^i=\epsilon^iz^i+(1-\epsilon^i)\tilde{z}^i$ for $i=1,...,m$; \\
		\STATE \qquad Update $\theta_C$ by minimizing $\mathcal{L}^C$ (Eq.~\ref{eq:loss_for_C}); \\
		\STATE \quad \textbf{end for} \\
		\STATE \quad Sample $\{x^i\}_{i=1}^{m}$ from $\mathcal{X}$; \\
		\STATE \quad Sample ${z'}^i=({z'}_{c}^i,{z'}_{s}^i)$ from $Q(Z|X=T(x^i))$ for $i=1,...,m$; \\
		\STATE \quad Sample $z^i=(z_{c}^i,z_{s}^i)$ from $Q(Z|X=x^i)$ for $i=1,...,m$; \\
		\STATE \quad Sample $z^j$ from $\{z^i\}_{i=1}^{m}$ for each $x^i$ to form negative paris; \\
		\STATE \quad Update $\theta_Q$ and $\theta_D$ by minimizing $\mathcal{L}^Q$ (Eq.~\ref{eq:loss_for_Q}) and $\mathcal{L}^D$ (Eq.~\ref{eq:loss_for_D}); \\
		\STATE \textbf{end while} \\
		\STATE \textbf{for} $i=1,...,N$ \textbf{do} \\
		\STATE \quad Sample $z^i=(z_{c}^i,z_{s}^i)$ from $Q(Z|X=x^i)$; \\
		\STATE \quad Compute cluster assignment $l^i=\text{argmax}(z_{c}^i)$; \\	
		\STATE \textbf{end for} \\
		\ENSURE  Cluster assignment $\{l^i\}_{i=1}^{N}$. \\
	\end{algorithmic}
\end{algorithm}

\section{Experiments}
\label{sec:experiment}
\subsection{Experimental Settings}

\subsubsection{Datasets.}
We evaluate the proposed DCCS on five commonly used datasets, including MNIST~\cite{lecun1998gradient}, Fashion-MNIST~\cite{xiao2017fashion}, CIFAR-10~\cite{krizhevsky2009learning}, STL-10~\cite{coates2011analysis} and ImageNet-10~\cite{chang2017deep}. The statistics of these datasets are described in Table~\ref{table:datasets}. As a widely adopted setting~\cite{xie2016unsupervised,chang2017deep,wu2019deep}, the training and test sets of these datasets are jointly utilized. For STL-10, the unlabelled subset is not used. For ImageNet-10, images are selected from the ILSVRC2012 1K dataset~\cite{deng2009imagenet} the same as in~\cite{chang2017deep} and resized to $96\times 96$ pixels. Similar to the IIC approach~\cite{ji2019invariant}, color images are converted to grayscale to discourage clustering based on trivial color cues.

\begin{table}[t]
	\centering 
	\caption{Statistics of the datasets}
	\begin{adjustbox}{max width=\linewidth}
	\begin{tabular}{l c c c}
		\shline
		Dataset & Images & Clusters & Image size \\
		\hline
		MNIST~\cite{lecun1998gradient} & 70000 & 10 & $28\times 28$ \\
		Fashion-MNIST~\cite{xiao2017fashion} & 70000 & 10 & $28\times 28$ \\
		CIFAR-10~\cite{krizhevsky2009learning} & 60000 & 10 & $32\times 32\times 3$ \\
		STL-10~\cite{coates2011analysis} & 13000 & 10 & $96\times 96\times 3$ \\
		ImageNet-10~\cite{chang2017deep} & 13000 & 10 & $96\times 96\times 3$ \\
		\shline
	\end{tabular}
	\end{adjustbox}
	\label{table:datasets}
\end{table}

\subsubsection{Evaluation metrics.}
Three widely used metrics are applied to evaluate the performance of the clustering methods, including unsupervised clustering accuracy~(ACC), normalized mutual information~(NMI), and adjusted rand index~(ARI)~\cite{wu2019deep}. For these metrics, a higher score implies better performance.

\subsubsection{Implementation details.}
The architectures of encoders are similar to~\cite{mukherjee2019clustergan,gulrajani2017improved} with a different number of layers and units being used for different sizes of images. The critic and discriminator are multi-layer perceptions. All the parameters are randomly initialized without pretraining. The Adam~\cite{kingma2014adam} optimizer with a learning rate of $10^{-4}$ and $\beta_1=0.5$, $\beta_2=0.9$ is used for optimization. The dimension of $Z_s$ is set to 50, and the dimension of $Z_c$ is set to the expected number of clusters. For other hyper-parameters, we set the standard deviation of prior Gaussian distribution $\sigma=0.1$, the gradient penalty coefficient $\lambda=10$, $\beta_{\text{MI}}=0.5$, $\beta_{\text{Adv}}=1$, the number of critic iterations per encoder iteration $n_{\text{critic}}=4$, and batch size $m=64$ for all datasets. Because $\beta_{\text{Aug}}$ is related to the datasets and generally the more complex the images are, the larger $\beta_{\text{Aug}}$ should be. We come up with an applicable way to set $\beta_{\text{Aug}}$ by visualizing the t-SNE figure of the encoded representation $Z$, \ie $\beta_{\text{Aug}}$ is gradually increased until the clusters visualized by t-SNE start to overlap. With this method, $\beta_{\text{Aug}}$ is set to 2 for MNIST and Fashion-MNIST, and set to 4 for other datasets. The data augmentation includes four commonly used approaches, \ie random cropping, random horizontal flipping, color jittering and channel shuffling~(which is used on the color images before graying). For more details about network architectures, data augmentation and hyper-parameters, please refer to the supplementary materials.

\subsection{Main Result}
\subsubsection{Quantitative comparison.} 
We first compare the proposed DCCS with several baseline methods as well as other state-of-the-art clustering approaches based on deep learning, as shown in Table~\ref{table:sota_res1} and Table~\ref{table:sota_res2}. DCCS outperforms all the other methods by large margins on Fashion-MNIST, CIFAR-10, STL-10 and ImageNet-10. For the ACC metric, DCCS is 2.0\%, 3.3\%, 3.7\% and 2.7\% higher than the second best methods on these four datasets, respectively. Although for MNIST, the clustering accuracy of DCCS is slightly lower~(\ie 0.3\%) than IIC~\cite{ji2019invariant}, DCCS significantly surpasses IIC on CIFAR-10 and STL-10.

\begin{table}[t]
    \centering
    \caption{Comparison with baseline and state-of-the-art methods on MNIST and Fashion-MNIST. The best three results of each metric are highlighted in \textbf{bold}. $\star$: Re-implemented results with the released code}
    \begin{adjustbox}{max width=\linewidth}
    \begin{tabular}{|c|ccc|ccc|}
		\hline
		\multirow{2}{*}{Method} & \multicolumn{3}{c|}{MNIST} & \multicolumn{3}{c|}{Fashion-MNIST}  \\ \cline{2-7}
		 & ACC & NMI & ARI    & ACC & NMI & ARI  \\ \hline
		K-means~\cite{wang2014optimized} & 0.572 & 0.500 & 0.365 & 0.474 & 0.512 & 0.348  \\ \hline
		SC~\cite{zelnik2005self} & 0.696 & 0.663 & 0.521 & 0.508 & 0.575 & -   \\ \hline
		AC~\cite{gowda1978agglomerative} & 0.695 & 0.609 & 0.481 & 0.500 & 0.564 & 0.371  \\ \hline
		NMF~\cite{cai2009locality} & 0.545 & 0.608 & 0.430 & 0.434 & 0.425 & -  \\ \hline
		DEC~\cite{xie2016unsupervised} & 0.843 & 0.772 & 0.741 & 0.590$^\star$ & 0.601$^\star$ & 0.446$^\star$  \\ \hline
		JULE~\cite{yang2016joint} & 0.964 & 0.913 & 0.927 & 0.563 & 0.608 & - \\ \hline
		VaDE~\cite{jiang2016variational} & 0.945 & 0.876 & - & 0.578 & 0.630 & -  \\ \hline
		DEPICT~\cite{ghasedi2017deep} & 0.965 & 0.917 & - & 0.392 & 0.392 & - \\ \hline
		IMSAT~\cite{hu2017learning} & \textbf{0.984} & \textbf{0.956$^\star$} & \textbf{0.965$^\star$} & \textbf{0.736$^\star$} & \textbf{0.696$^\star$} & \textbf{0.609$^\star$}\\ \hline
		DAC~\cite{chang2017deep} & 0.978 & 0.935 & 0.949 & 0.615$^\star$ & 0.632$^\star$ & 0.502$^\star$  \\ \hline
		SpectralNet~\cite{shaham2018spectralnet} & 0.971 & 0.924 & 0.936$^\star$ & 0.533$^\star$ & 0.552$^\star$ & - \\ \hline
		ClusterGAN~\cite{mukherjee2019clustergan} & 0.950 & 0.890 & 0.890 & 0.630 & 0.640 & 0.500 \\ \hline
		DLS-Clustering~\cite{ding2019clustering} & 0.975 & 0.936 & - & 0.693 & 0.669 & -  \\ \hline
		DualAE~\cite{yang2019deep} & 0.978 & 0.941 & - & 0.662 & 0.645 & - \\ \hline
		RTM~\cite{nina2019decoder} & 0.968 & 0.933 & 0.932 & 0.710 & 0.685 & 0.578  \\ \hline
		NCSC~\cite{zhang2019neural} & 0.941 & 0.861 & 0.875 & \textbf{0.721} & \textbf{0.686} & \textbf{0.592} \\ \hline
		IIC~\cite{ji2019invariant} & \textbf{0.992} & \textbf{0.978$^\star$} & \textbf{0.983$^\star$} & 0.657$^\star$ & 0.637$^\star$ & 0.523$^\star$ \\ \hline
		DCCS~(Proposed)    & \textbf{0.989}       &   \textbf{0.970}         & \textbf{0.976}    & \textbf{0.756}     & \textbf{0.704}     & \textbf{0.623}  \\ \hline
	\end{tabular}
    \end{adjustbox}
    \label{table:sota_res1}
\end{table}

\begin{table}[t]
    \centering
    \caption{Comparison with baseline and state-of-the-art methods on CIFAR-10, STL-10 and ImageNet-10. The best three results of each metric are highlighted in \textbf{bold}. $\star$: Re-implemented results with the released code. $\dagger$: The results are evaluated on STL-10 without using the unlabelled data subset}
    \begin{adjustbox}{max width=\linewidth}
    \begin{tabular}{|c|ccc|ccc|ccc|}
		\hline
		\multirow{2}{*}{Method} & \multicolumn{3}{c|}{CIFAR-10} & \multicolumn{3}{c|}{STL-10} & \multicolumn{3}{c|}{ImageNet-10} \\ \cline{2-10}
		 & ACC & NMI & ARI   & ACC & NMI & ARI   & ACC & NMI & ARI \\ \hline
		K-means~\cite{wang2014optimized} & 0.229 & 0.087 & 0.049 & 0.192  & 0.125  & 0.061  & 0.241  & 0.119  & 0.057 \\ \hline
		SC~\cite{zelnik2005self} & 0.247 & 0.103 & 0.085 & 0.159  & 0.098  & 0.048  & 0.274  & 0.151  & 0.076  \\ \hline
		AC~\cite{gowda1978agglomerative} & 0.228 & 0.105 & 0.065 & 0.332  & 0.239  & 0.140  & 0.242  & 0.138  & 0.067 \\ \hline
		NMF~\cite{cai2009locality} & 0.190 & 0.081 & 0.034 & 0.180  & 0.096  & 0.046  & 0.230  & 0.132  & 0.065  \\ \hline
		AE~\cite{bengio2007greedy} & 0.314 & 0.239 & 0.169 & 0.303 & 0.250 & 0.161  & 0.317 & 0.210 & 0.152\\ \hline
		GAN~\cite{radford2015unsupervised} & 0.315 & 0.265 & 0.176 & 0.298 & 0.210 & 0.139  & 0.346 & 0.225 & 0.157\\ \hline
		VAE~\cite{kingma2013auto} & 0.291 & 0.245 & 0.167 & 0.282 & 0.200 & 0.146  & 0.334 & 0.193 & 0.168\\ \hline
		DEC~\cite{xie2016unsupervised} & 0.301 & 0.257 & 0.161 & 0.359  & 0.276  & 0.186  & 0.381  & 0.282  & 0.203  \\ \hline
		JULE~\cite{yang2016joint} & 0.272 & 0.192 & 0.138 & 0.277  & 0.182  & 0.164  & 0.300  & 0.175  & 0.138 \\ \hline
		DAC~\cite{chang2017deep} & 0.522 & 0.396 & 0.306 & 0.470  & 0.366  & 0.257  & \textbf{0.527}  & \textbf{0.394}  & \textbf{0.302}  \\ \hline
		IIC~\cite{ji2019invariant} & \textbf{0.617} & \textbf{0.513$^\star$} & \textbf{0.411$^\star$} & \textbf{0.499$^\dagger$} & \textbf{0.431$^{\star\dagger}$}  & \textbf{0.295$^{\star\dagger}$}  & -  & -  & - \\ \hline
		DCCM~\cite{wu2019deep} & \textbf{0.623} & \textbf{0.496} & \textbf{0.408} & \textbf{0.482}  & \textbf{0.376}  & \textbf{0.262}  & \textbf{0.710}  & \textbf{0.608}  & \textbf{0.555} \\ \hline
		DCCS~(Proposed)     &  \textbf{0.656}          & \textbf{0.569} &  \textbf{0.469}      &     \textbf{0.536}       & \textbf{0.490} &   \textbf{0.362}     &      \textbf{0.737}      &  \textbf{0.640} &  \textbf{0.560}   \\ \hline
	\end{tabular}
    \end{adjustbox}
    \label{table:sota_res2}
\end{table}

\subsubsection{Training progress.} 
\label{sec:training_progress}
\begin{figure}[t]
    \centering
    \includegraphics[width=0.75\linewidth]{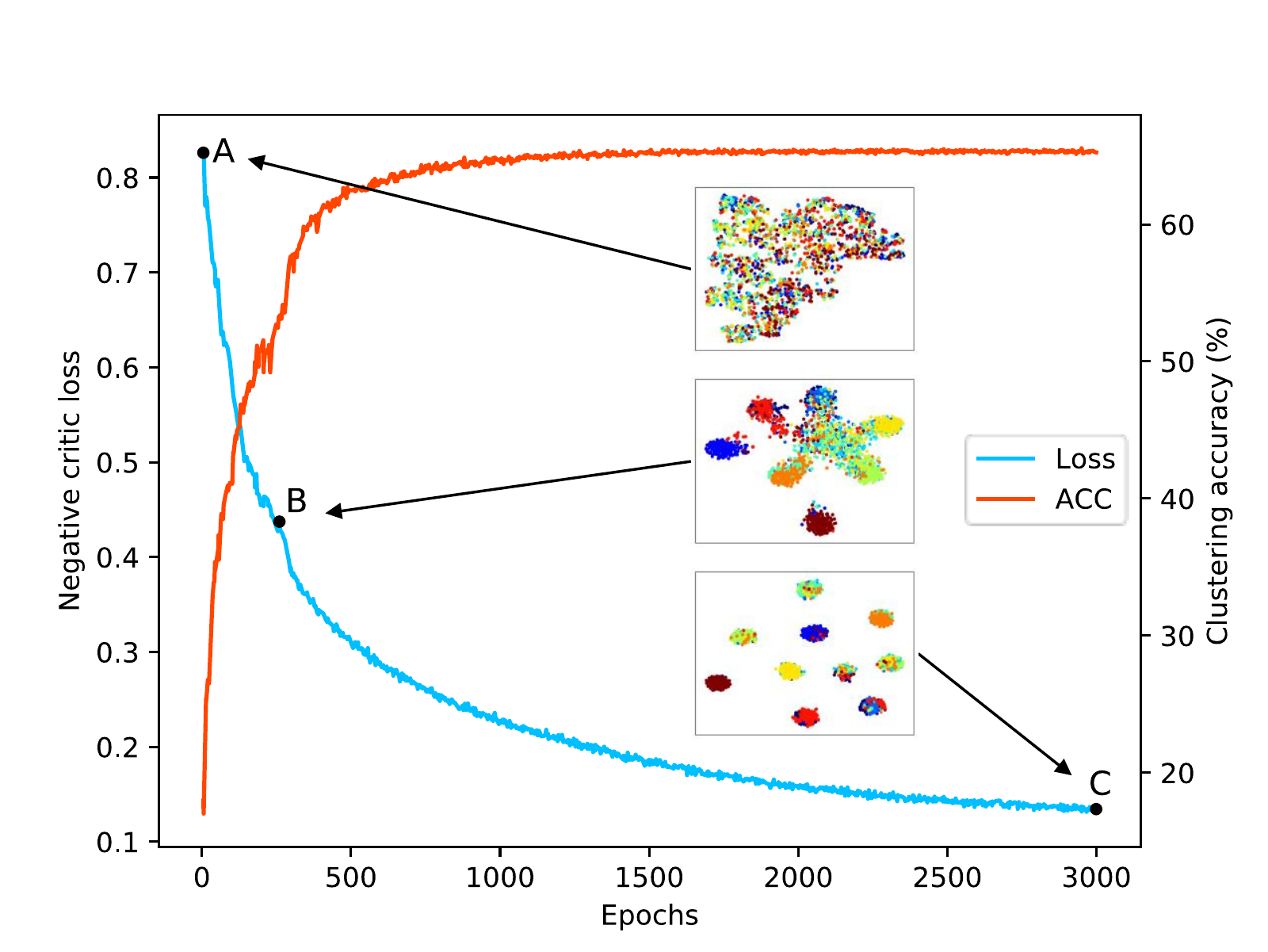}
    \caption{Training curves of the negative critic loss and the clustering accuracy on CIFAR-10. The t-SNE visualizations of the latent representations $Z$ for different stages are also displayed. The color of the points in the t-SNE visualizations refers to the ground truth category}
    \label{fig:training_curves}
\end{figure}

The training progress of the proposed DCCS is monitored by minimizing the Wasserstein distance $W(Q_Z,P_Z)$, which can be estimated by the negative critic loss $-\mathcal{L}^C$. As plotted in Fig.~\ref{fig:training_curves}, the critic loss stably converges and it correlates well with the clustering accuracy, demonstrating a robust training progress. The visualizations of the latent representations with t-SNE at three different stages are also displayed in Fig.~\ref{fig:training_curves}. From stage A to C, the latent representations gradually cluster together while the critic loss decreases steadily.

\subsubsection{Qualitative analysis.} 
\begin{figure}[t]
	\centering
	  \subfloat[]{\label{fig:cluster_imgs_mnist}%
		\includegraphics[width=.45\linewidth]{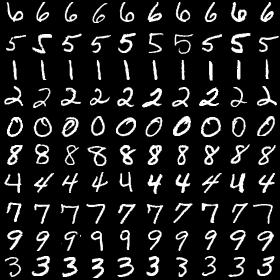}}
	  \quad
	  \subfloat[]{\label{fig:cluster_imgs_imagenet10}%
		\includegraphics[width=.45\linewidth]{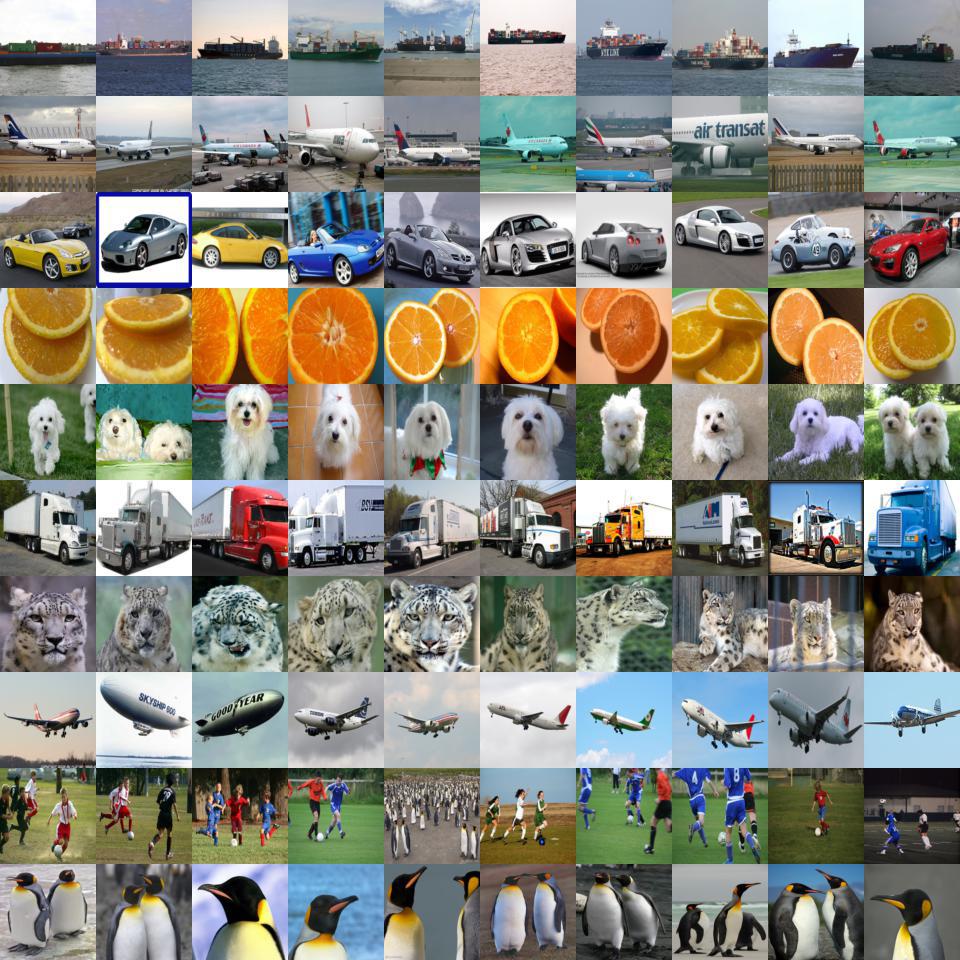}}
	  \caption{Clustering images from MNIST (a) and ImageNet-10 (b). Each row contains the images with the highest probability to belong to the respective cluster}
	  \label{fig:qualitative_analysis}
\end{figure}

Fig.~\ref{fig:qualitative_analysis} shows images with top 10 predicted probabilities from each cluster in MNIST and ImageNet-10. Each row corresponds to a cluster and the images from left to right are sorted in a descending order based on their probabilities. In each row of Fig.~\ref{fig:cluster_imgs_mnist}, the same digits are written in different ways, indicating that $Z_c$ contains well disentangled category information. For ImageNet-10 in Fig.~\ref{fig:cluster_imgs_imagenet10}, most objects are well clustered and the major confusion is for the airships and airplanes in the sky due to their similar shapes and backgrounds~(Row 8). A possible solution is overclustering, \ie more number of clusters than expected, which requires investigation in future work.

\begin{figure}[!th]
	\centering
	  \subfloat[]{\label{fig:beta_aug_fashion_mnist}%
		\includegraphics[width=.45\linewidth]{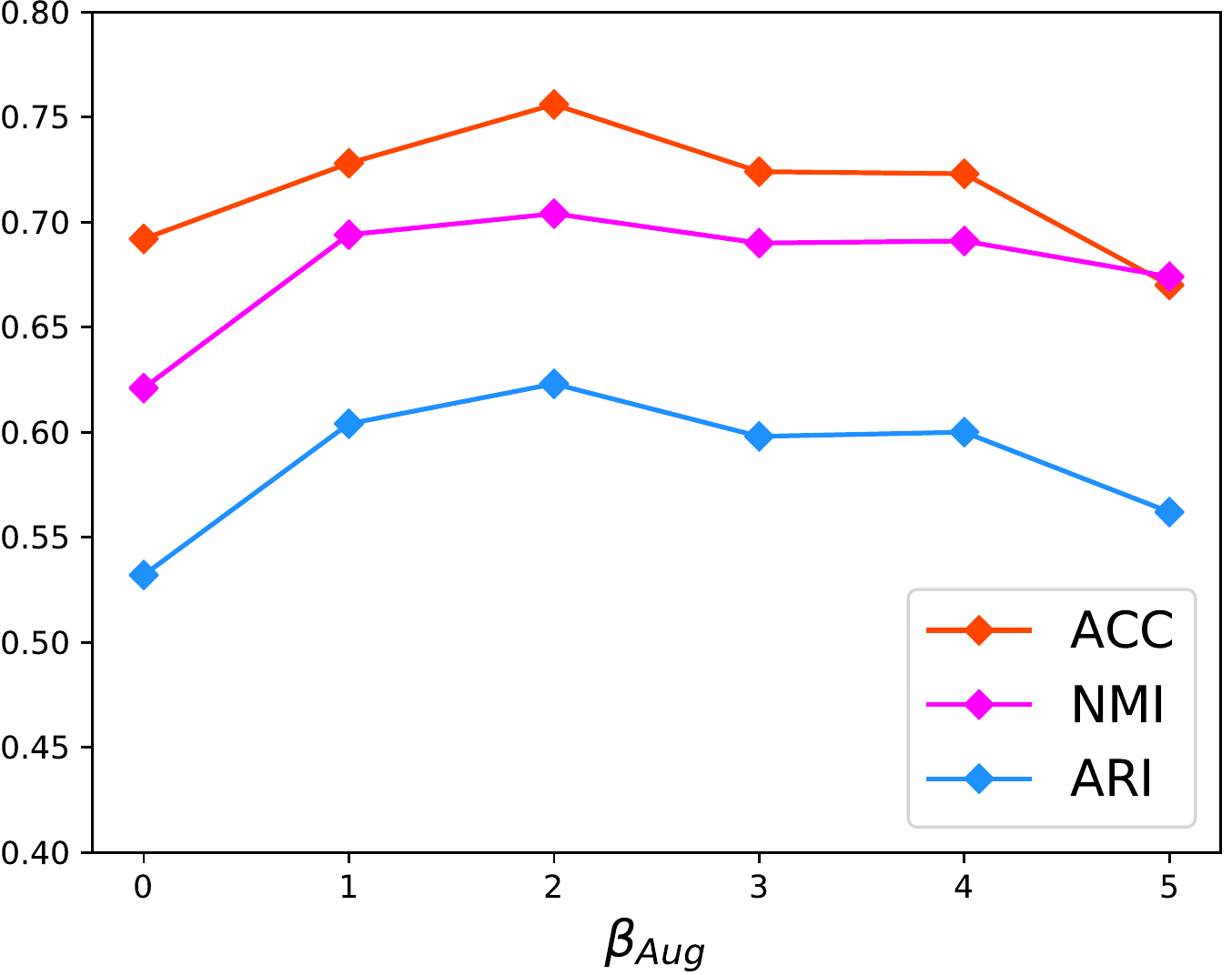}}
	  \quad
	  \subfloat[]{\label{fig:beta_aug_fashion_cifar10}%
		\includegraphics[width=.45\linewidth]{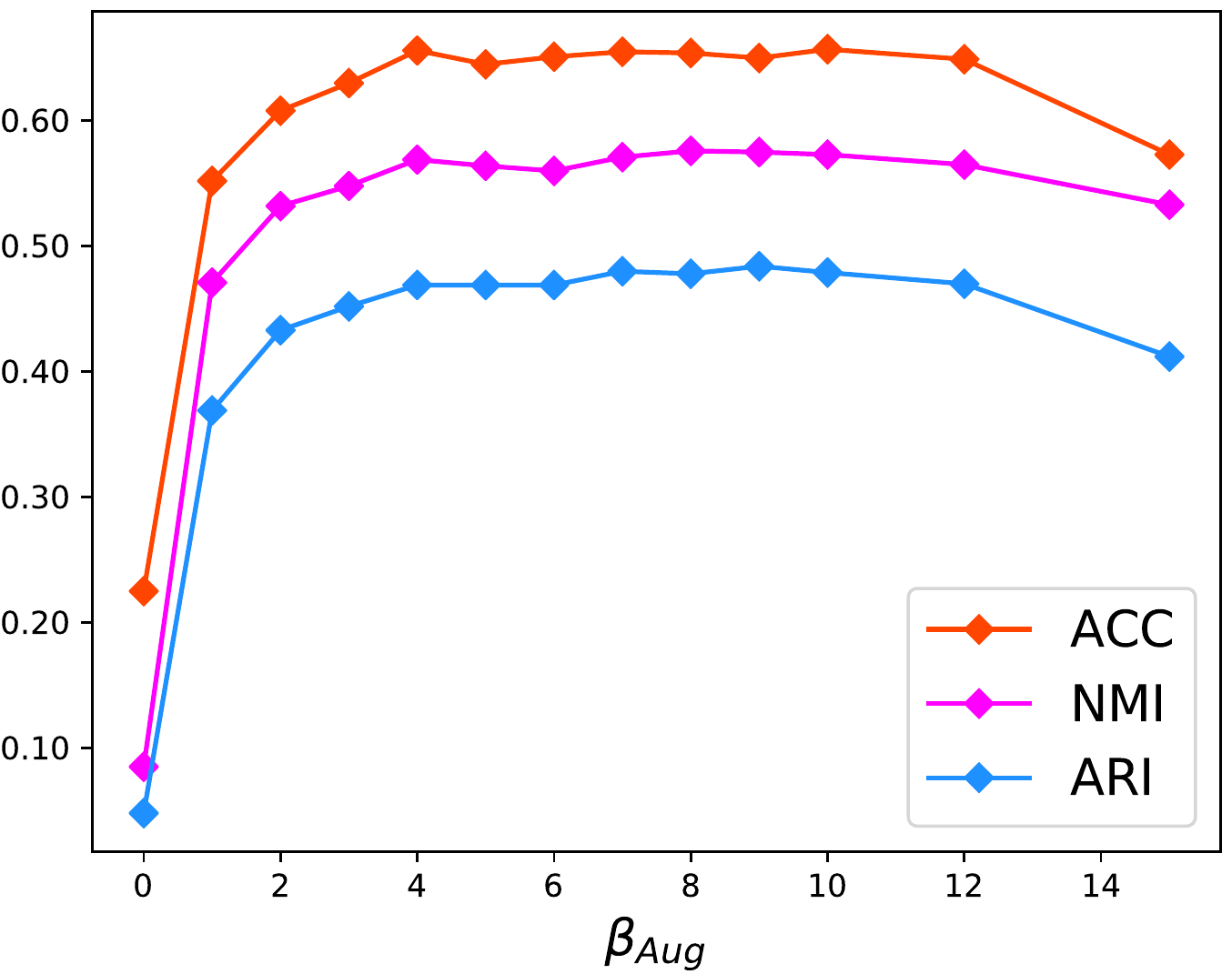}}
	  \caption{The impact of $\beta_{\text{Aug}}$ on Fashion-MNIST (a) and CIFAR-10 (b)}
	  \label{fig:vary_beta_aug}
\end{figure}

\subsection{Ablation Study}
\label{sec:ablation_study}
\subsubsection{Cluster assignment w/o K-means.}
As stated in Section~\ref{sec:match_to_prior}, by imposing the prior distribution, the latent category representation $Z_c$ can be directly used as the cluster assignment. Table~\ref{table:encoder_vs_kmeans} compares the results of several ways to obtain the cluster assignment with the same encoder. We can see that using $Z_c$ with or without K-means has similar performance, indicating that $Z_c$ is discriminative enough to be used as the cluster assignment directly. Additional experiments on each part of $Z$ show that applying K-means on $Z_c$ can yield similar performance as on $Z$, while the performance of applying K-means on $Z_s$ is much worse. It demonstrates that the categorical cluster information and the style information are well disentangled as expected.

\begin{table}[t]
    \centering
    \caption{Evaluation of different ways for the cluster assignment}
    \begin{adjustbox}{max width=\linewidth}
    \begin{tabular}{|c|ccc|ccc|ccc|ccc|}
		\hline
		\multirow{2}{*}{Method} & \multicolumn{3}{c|}{MNIST} & \multicolumn{3}{c|}{Fashion-MNIST} & \multicolumn{3}{c|}{CIFAR-10} & \multicolumn{3}{c|}{STL-10}  \\ \cline{2-13}
		 & ACC & NMI & ARI    & ACC & NMI & ARI   & ACC & NMI & ARI   & ACC & NMI & ARI  \\ \hline
		Argmax over $Z_c$    & \textbf{0.9891}       &   \textbf{0.9696}         & \textbf{0.9758}    & \textbf{0.7564}     & 0.7042     & \textbf{0.6225}        &  \textbf{0.6556}          & \textbf{0.5692} &  0.4685      &     \textbf{0.5357}       & \textbf{0.4898} &   \textbf{0.3617}    \\ \hline
		K-means on $Z$ & \textbf{0.9891} & 0.9694 & 0.9757 & \textbf{0.7564} & \textbf{0.7043} & 0.6224 & 0.6513 & 0.5588 & \textbf{0.4721} & 0.5337  & 0.4888  & 0.3599  \\ \hline
		K-means on $Z_c$ & \textbf{0.9891} & \textbf{0.9696} & 0.9757 & 0.7563 & 0.7042 & 0.6223 & 0.6513 & 0.5587 & \textbf{0.4721} & 0.5340  & 0.4889  & 0.3602  \\ \hline
		K-means on $Z_s$ & 0.5164 & 0.4722 & 0.3571 & 0.4981 & 0.4946 & 0.3460 & 0.2940 & 0.1192 & 0.0713 & 0.4422  & 0.4241  & 0.2658  \\ \hline
	\end{tabular}
    \end{adjustbox}
    \label{table:encoder_vs_kmeans}
\end{table}

\subsubsection{Ablation study on the losses.}
The effectiveness of the losses is evaluated in Table~\ref{table:ablation_study}. M1 is the baseline method, \ie the only constraint applied to the network is the prior distribution. This constraint is always necessary to ensure that the category representation can be directly used as the cluster assignment. By adding the mutual information maximization in M2 or the category-style information disentanglement in M3, the clustering performance achieves significant gains. The results of M4 demonstrate that combining all three losses can further improve the clustering performance. Note that large improvement with data augmentation for CIFAR-10 is due to that the images in CIFAR-10 have considerable intra-class variability, therefore disentangling the category-style information can improve the clustering performance by a large margin. 

\begin{table}[t]
    \centering
    \caption{Ablation study of DCCS on Fashion-MNIST and CIFAR-10}
    \begin{adjustbox}{max width=\linewidth}
    \begin{tabular}{|c|ccc|ccc|ccc|}
		\hline
		\multirow{2}{*}{Method} & \multicolumn{3}{|c|}{Loss} & \multicolumn{3}{c|}{Fashion-MNIST} & \multicolumn{3}{c|}{CIFAR-10} \\ \cline{2-10}
		& $\mathcal{L}_{\text{Adv}}$ & $\mathcal{L}_{\text{MI}}$ & $\mathcal{L}_{\text{Aug}}$ & ACC & NMI & ARI    & ACC & NMI & ARI  \\ \hline
		M1 & \checkmark & & & 0.618 & 0.551 & 0.435 & 0.213 & 0.076 & 0.040 \\ \hline
		M2 & \checkmark & \checkmark & & 0.692 & 0.621 & 0.532 & 0.225 & 0.085 & 0.048 \\ \hline
		M3 & \checkmark & & \checkmark & 0.725 & 0.694 & 0.605 & 0.645 & 0.557 & 0.463 \\ \hline
		M4 & \checkmark & \checkmark & \checkmark & \textbf{0.756} & \textbf{0.704} & \textbf{0.623} & \textbf{0.656} & \textbf{0.569} & \textbf{0.469} \\ \hline
	\end{tabular}
    \end{adjustbox}
    \label{table:ablation_study}
\end{table}

\subsubsection{Impact of $\beta_{\text{Aug}}$.}
The clustering performance with different $\beta_{\text{Aug}}$, which is the weight of the data augmentation loss in Eq.~\ref{eq:loss_for_Q},  is displayed in Fig.~\ref{fig:vary_beta_aug}. For Fashion-MNIST, the performance drops when $\beta_{\text{Aug}}$ is either too small or too large because a small $\beta_{\text{Aug}}$ cannot disentangle the style information enough, and a large $\beta_{\text{Aug}}$ may lead the clusters to overlap. For CIFAR-10, the clustering performance is relatively stable with large $\beta_{\text{Aug}}$. As previously stated, the biggest $\beta_{\text{Aug}}$ without overlapping clusters in the t-SNE visualization of the encoded representation $Z$ is selected~(the visualization of t-SNE can be found in the supplementary materials).

\subsubsection{Impact of $Z_s$.} 
\begin{figure}[t]
	\centering
	  \subfloat[]{\label{fig:dim_zs_fashion_mnist}%
		\includegraphics[width=.45\linewidth]{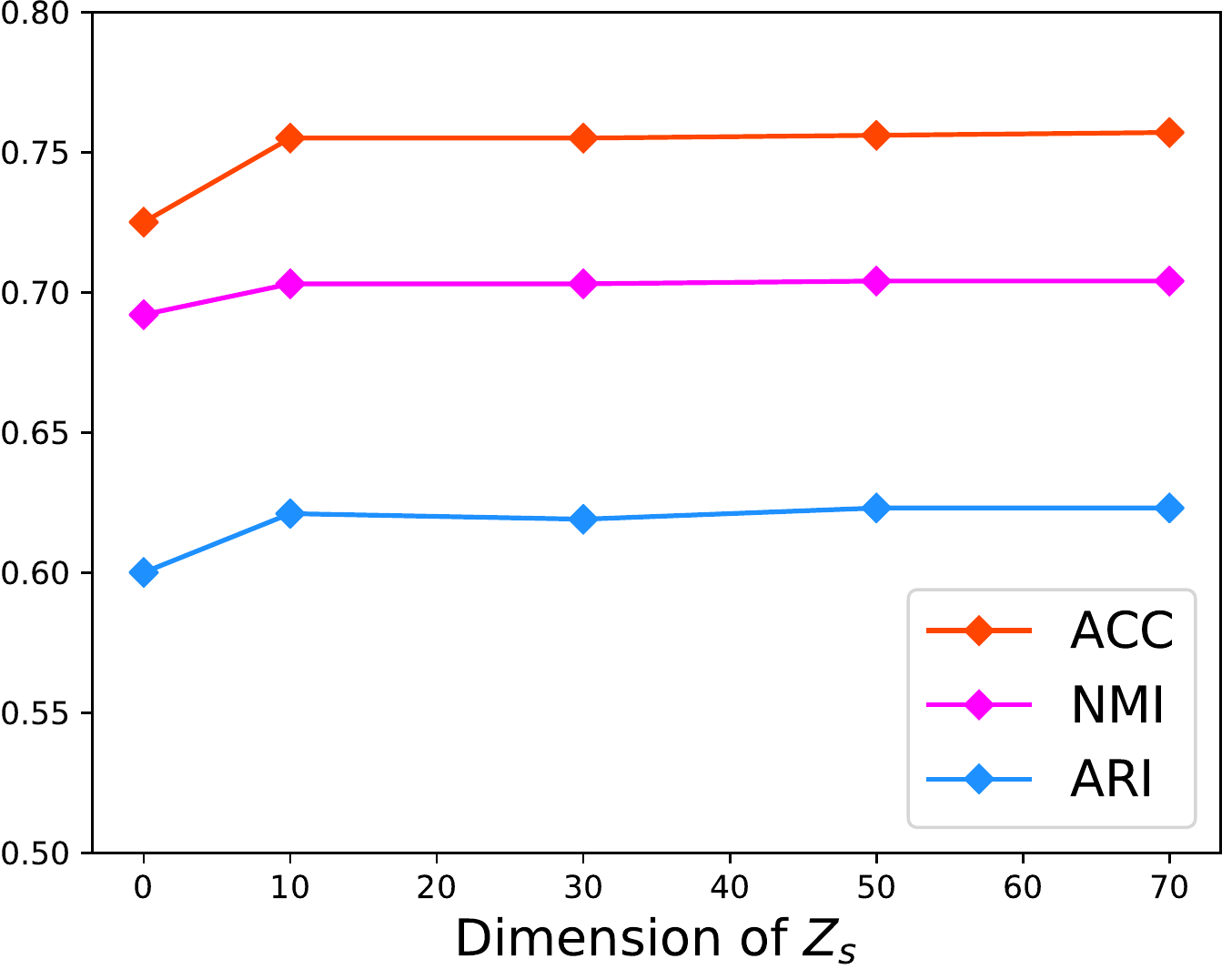}}
	  \quad
	  \subfloat[]{\label{fig:dim_zs_fashion_cifar10}%
		\includegraphics[width=.45\linewidth]{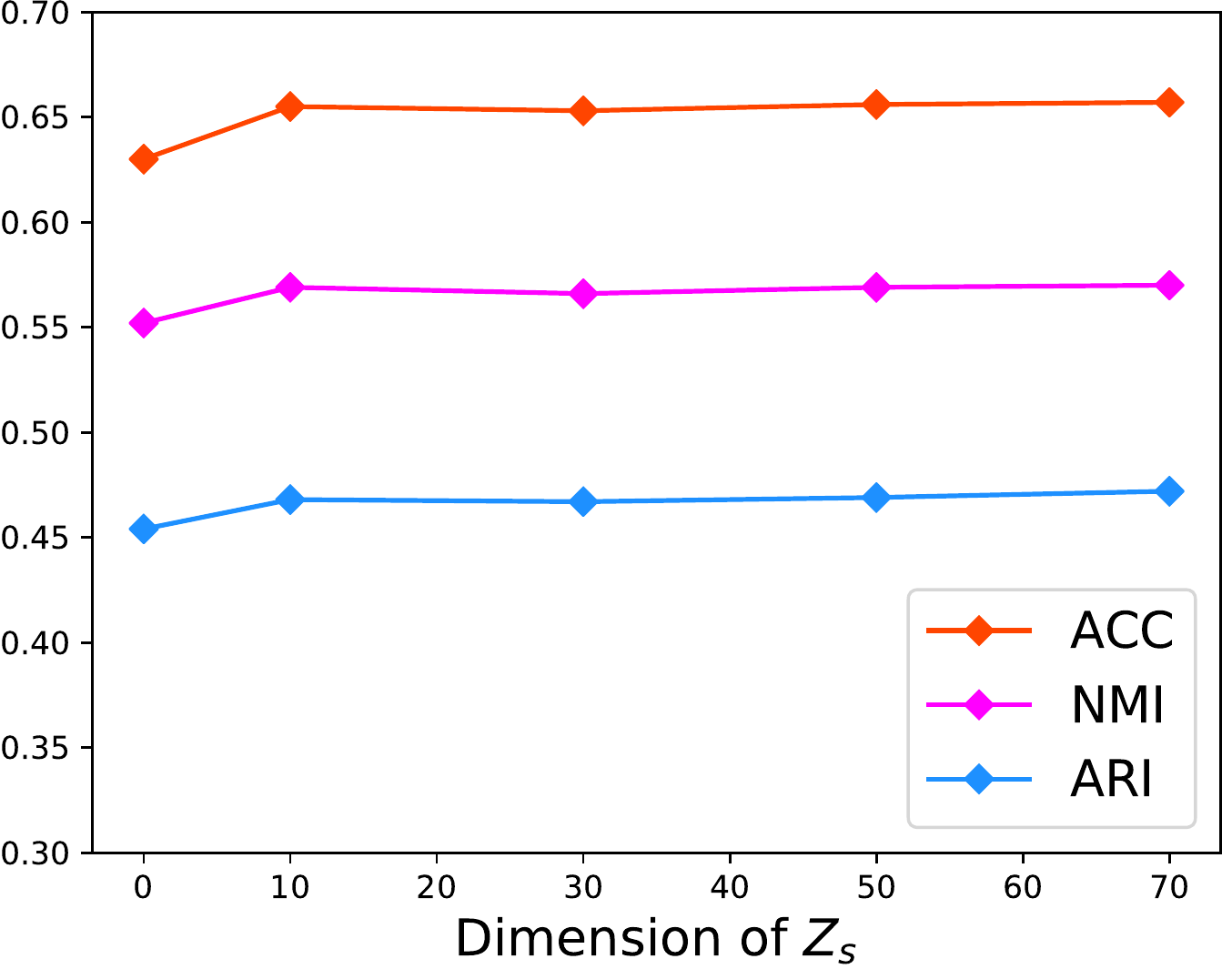}}
	  \caption{The impact of $Z_s$ on Fashion-MNIST (a) and CIFAR-10 (b)}
	  \label{fig:vary_dim_zs}
\end{figure}

As shown in Fig.~\ref{fig:vary_dim_zs}, varying the dimension of $Z_s$ from 10 to 70 does not affect the clustering performance much. However, when the dimension of $Z_s$ is 0, \ie the latent representation only contains the category part, the performance drops a lot, demonstrating the necessity of the style representation. The reason is that for the mutual information maximization, the category representation alone is not enough to describe the difference among images belonging to the same cluster.

\section{Conclusions}
In this work, we proposed a novel unsupervised deep image clustering framework with three regularization strategies. First, mutual information maximization was applied to retain essential information and avoid arbitrary representations. Furthermore, data augmentation was employed to disentangle the category representation from style information. Finally, a prior distribution was imposed to prevent degenerate solutions and avoid the usage of additional clustering so that the category representation could be used directly as the cluster assignment. Ablation studies demonstrated the effectiveness of each component and the extensive comparison experiments on five datasets showed that the proposed approach outperformed other state-of-the-art methods.

\section*{Acknowledgements}
This work was supported by National Key Research and Development Program of China (No. 2018AAA0100100), National Natural Science Foundation of China (61702095), the Key Area Research and Development Program of Guangdong Province, China (No. 2018B010111001), National Key Research and Development Project (2018YFC2000702), Science and Technology Program of Shenzhen, China (No. ZDSYS201802021814180) and the Big Data Computing Center of Southeast University.

\clearpage
%
%
\bibliographystyle{splncs04}
\bibliography{main}

\clearpage

\appendix

\begin{center}
\Large{\textbf{Supplementary Materials}}
\end{center}

\section{Network Architectures}
\begin{figure}[!th]
    \centering
    \includegraphics[width=0.6\linewidth]{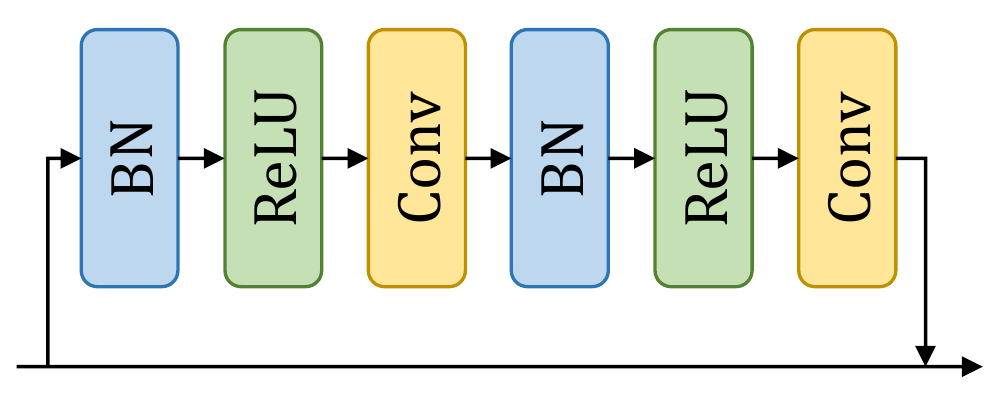}
    \caption{ResBlock architecture. The kernel size of the convolutional layer is $3\times 3$. $2\times 2$ average pooling is employed for downsampling after the second convolution, while the nearest-neighbor upsampling is applied for upsampling before the first convolution}
    \label{fig:res_block}
\end{figure}

The architectures of the encoders with respect to different size of images are described in Table~\ref{table:encoder28}, Table~\ref{table:encoder32} and Table~\ref{table:encoder96}. The ResBlocks used in the encoders are presented in Fig.~\ref{fig:res_block}. Table~\ref{table:critic} and Table~\ref{table:discriminator} show the architecture of the critic and the discriminator, respectively. The critic and the discriminator both consist of three fully-connected layers. The discriminator shares some layers with the encoder to reduce computations, \ie the input $X$ in Table~\ref{table:discriminator} is a flatten vector created by the encoder. For the encoder in Table~\ref{table:encoder28}, $X$ is the output of the last convolutional layer, while for the encoders in Table~\ref{table:encoder32} and Table~\ref{table:encoder96}, $X$ is the output of the second last ResBlock. The slopes of lReLU functions for all architectures are set to 0.2.

\section{Data Augmentation}
The data augmentation adopted in DCCS includes four commonly used approaches:
\begin{enumerate}[(1)]
	\item Random cropping: randomly crop a rectangular region whose aspect ratio and area are randomly sampled in the range of $[3/4, 4/3]$ and $[40\%, 100\%]$, respectively, and then resize the cropped region to the original image size.
	\item Random horizontal flipping: flip the image horizontally with 50\% probability.
	\item Color jittering: scale brightness, contrast and saturation with coefficients uniformly drawn from $[0.6, 1.4]$, while scale hue with coefficients uniformly drawn from $[0.875, 1.125]$.
	\item Channel shuffling: randomly shuffle the RGB channels of the image.
\end{enumerate}
Random cropping and color jittering are employed for all datasets. Following~\cite{ji2019invariant}, random horizontal flipping is used for all datasets except MNIST due to the direction sensitive nature of the digits. Channel shuffling is applied to color images before graying. Note that channel shuffling can also change the brightness of the grayscale images because the RGB channels are summed with different weights for graying.

\begin{table}[t]
    \centering
    \caption{The encoder architecture for MNIST and Fashion-MNIST, similarly as the architecture used in~\cite{mukherjee2019clustergan}}
    \begin{adjustbox}{max width=\linewidth}
    \begin{tabular}{c}
		\hline 
		\hline 
		Input $X\in \mathbb{R}^{28\times 28}$\\ \hline
		$4\times 4$, stride=2 conv, BN 64 lReLU \\ \hline
		$4\times 4$, stride=2 conv, BN 128 lReLU \\ \hline
		Dense, BN 1024 lReLU \\ \hline
		Dense softmax for $Z_c$ \\ 
		Dense linear for $Z_s$ \\ \hline
		\hline
	\end{tabular}
    \end{adjustbox}
    \label{table:encoder28}
\end{table}

\begin{table}[t]
    \begin{minipage}{0.47\linewidth}
    \centering
    \caption{The encoder architecture for CIFAR-10, similarly as the architecture used in~\cite{gulrajani2017improved} with images converted to grayscale}
    \begin{adjustbox}{max width=\linewidth}
    \begin{tabular}{c}
		\hline 
		\hline 
		Input $X\in \mathbb{R}^{32\times 32}$\\ \hline
		ResBlock down 128 \\ \hline
		ResBlock down 256 \\ \hline
		ResBlock down 512 \\ \hline
		ResBlock 512 \\ \hline
		BN, ReLU, global average pooling \\ \hline
		Dense softmax for $Z_c$ \\ 
		Dense linear for $Z_s$ \\ \hline
		\hline
	\end{tabular}
    \end{adjustbox}
    \label{table:encoder32}
    \end{minipage}
    \hfill
    \begin{minipage}{0.47\linewidth}
    \centering
    \caption{The encoder architecture for STL-10 and ImageNet-10, similarly as the architecture used in~\cite{gulrajani2017improved} with images converted to grayscale}
    \begin{adjustbox}{max width=\linewidth}
    \begin{tabular}{c}
		\hline 
		\hline 
		Input $X\in \mathbb{R}^{96\times 96}$\\ \hline
		ResBlock down 64 \\ \hline
		ResBlock down 128 \\ \hline
		ResBlock down 256 \\ \hline
		ResBlock down 512 \\ \hline
		ResBlock 512 \\ \hline
		BN, ReLU, global average pooling \\ \hline
		Dense softmax for $Z_c$ \\ 
		Dense linear for $Z_s$ \\ \hline
		\hline
	\end{tabular}
    \end{adjustbox}
    \label{table:encoder96}
    \end{minipage}
\end{table}

\begin{table}[t]
    \begin{minipage}{0.45\linewidth}
    \centering
    \caption{The critic architecture}
    \begin{adjustbox}{max width=\linewidth}
    \begin{tabular}{c}
		\hline 
		\hline 
		Input $Z=(Z_c,Z_s)$\\ \hline
		Dense, 1024 lReLU \\ \hline
		Dense, 512 lReLU \\ \hline
		Dense, 1 linear \\ \hline
		\hline
	\end{tabular}
    \end{adjustbox}
    \label{table:critic}
    \end{minipage}
    \hfill
    \begin{minipage}{0.5\linewidth}
    \centering
    \caption{The discriminator architecture}
    \begin{adjustbox}{max width=\linewidth}
    \begin{tabular}{c}
		\hline 
		\hline 
		Input $(X,Z)$\\ \hline
		Dense, 1024 lReLU \\ \hline
		Dense, 512 lReLU \\ \hline
		Dense, 1 sigmoid \\ \hline
		\hline
	\end{tabular}
    \end{adjustbox}
    \label{table:discriminator}
    \end{minipage}
\end{table}

\begin{figure}[!th]
	\centering
	  \subfloat[]{\label{fig:tsne_fashion_mnist_aug2}%
		\includegraphics[width=.25\linewidth]{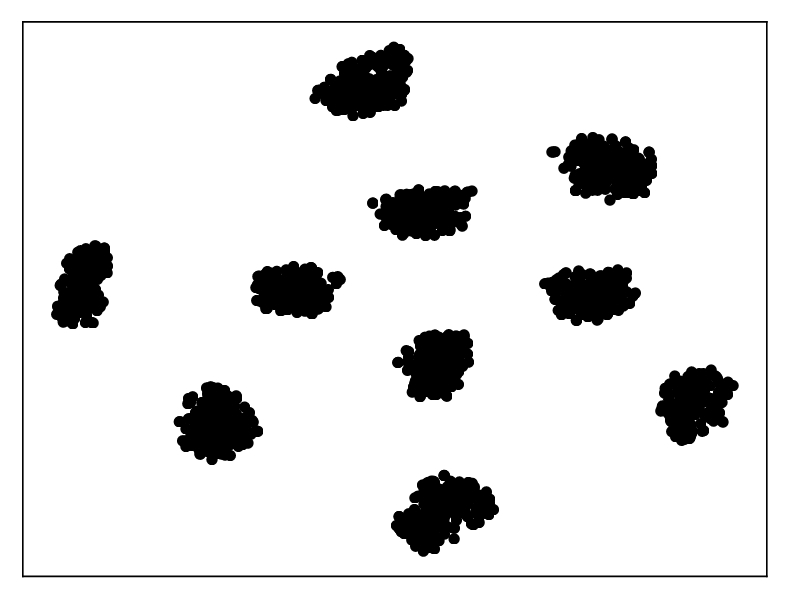}}
	  \subfloat[]{\label{fig:tsne_fashion_mnist_aug3}%
		\includegraphics[width=.25\linewidth]{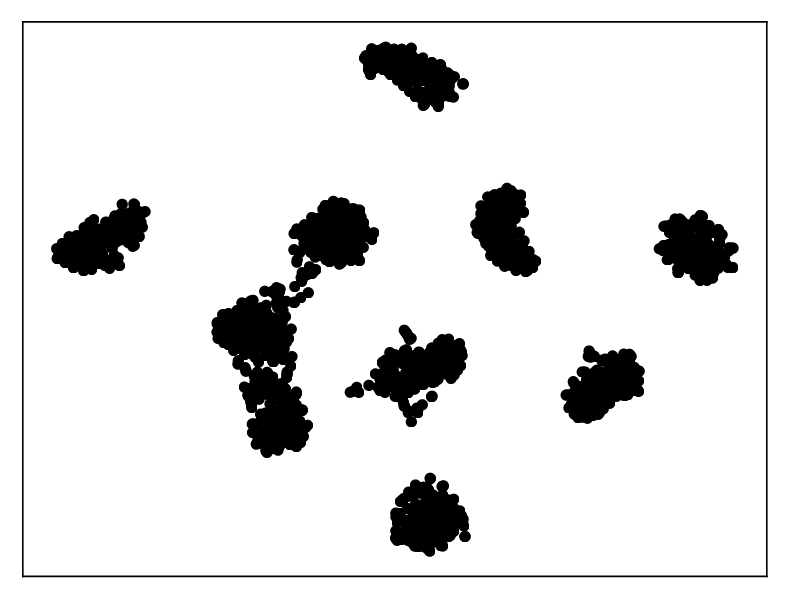}}
	  \subfloat[]{\label{fig:tsne_cifar10_aug4}%
		\includegraphics[width=.25\linewidth]{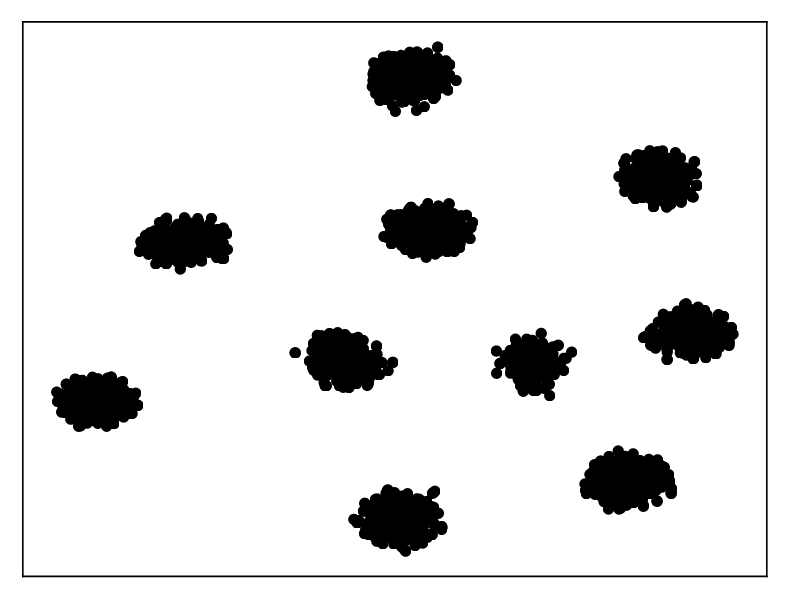}}
	  \subfloat[]{\label{fig:tsne_cifar10_aug5}%
		\includegraphics[width=.25\linewidth]{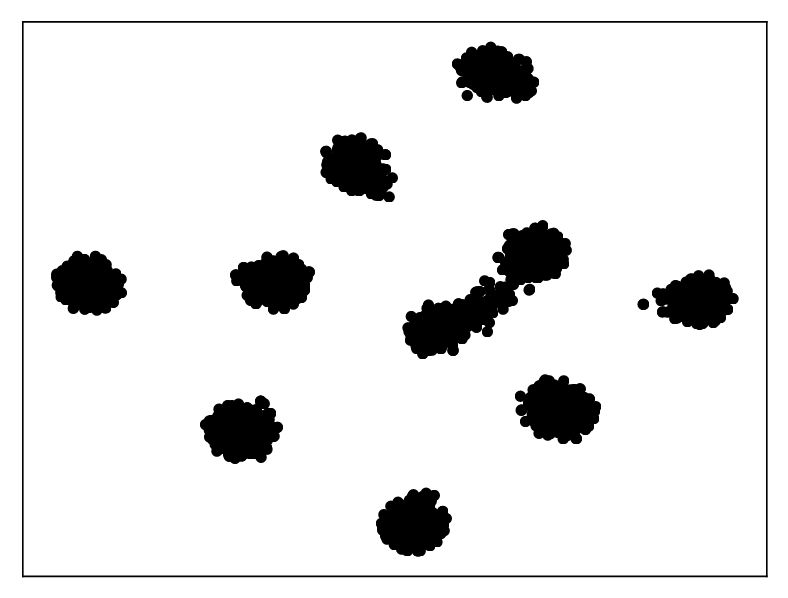}}
	  \caption{The t-SNE visualizations of the latent representations, including Fashion-MNIST with $\beta_{\text{Aug}}=2$ (a), Fashion-MNIST with $\beta_{\text{Aug}}=3$ (b), CIFAR-10 with $\beta_{\text{Aug}}=4$ (c) and CIFAR-10 with $\beta_{\text{Aug}}=5$ (d). Note that the visualization is completely based on the latent representation without any usage of the ground truth label}
	  \label{fig:set_beta_aug}
\end{figure}

\section{$\beta_{\text{Aug}}$ Configuration}
As previously stated, a small $\beta_{\text{Aug}}$ cannot disentangle the style information well, while a large $\beta_{\text{Aug}}$ may lead the clusters to overlap by generating high confidence of the overlapping part of two clusters. Therefore, we propose an applicable way for $\beta_{\text{Aug}}$ configuration by visualizing the t-SNE figure of the latent representation. As shown in Fig.~\ref{fig:tsne_fashion_mnist_aug2} and Fig.~\ref{fig:tsne_cifar10_aug4}, with $\beta_{\text{Aug}}$ being set to 2 for Fashion-MNIST and 4 for CIFAR-10, the clusters are well separated. However, the clusters start to overlap after increasing $\beta_{\text{Aug}}$ to 3 for Fashion-MNIST~(Fig.~\ref{fig:tsne_fashion_mnist_aug3}) or 5 for CIFAR-10~(Fig.~\ref{fig:tsne_cifar10_aug5}). Experiments show that using the biggest $\beta_{\text{Aug}}$ without overlapping clusters in the t-SNE visualization can always yield decent clustering performance.

\section{Discriminator vs. Decoder}
The proposed DCCS adopts a discriminator to maximize the mutual information $I(X,Z)$ between the input image $X$ and its latent representation $Z$ to avoid learning arbitrary representations. Autoencoder is another popular approach to embed the image information into the latent representation. The discriminator in the proposed framework could be replaced by a decoder with the mutual information loss being replaced by the reconstruction loss. The performance of these two approaches is compared in Table~\ref{table:discriminator_vs_decoder}. The decoder architectures for Fashion-MNIST and CIFAR-10 are described in Table~\ref{table:decoder28} and Table~\ref{table:decoder32}, respectively. The weight of the reconstruction loss is set to 5 for its best performance. The results show that the reconstruction strategy delivers inferior performance, suggesting that the representations learned by the decoder based DCCS may contain generative information which is irrelevant for clustering.

\begin{table}[!th]
    \centering
    \caption{Comparison of different ways to avoid learning arbitrary representations}
    \begin{adjustbox}{max width=\linewidth}
    \begin{tabular}{|c|ccc|ccc|}
		\hline
		\multirow{2}{*}{Method} & \multicolumn{3}{c|}{Fashion-MNIST} & \multicolumn{3}{c|}{CIFAR-10} \\ \cline{2-7}
		& ACC & NMI & ARI    & ACC & NMI & ARI  \\ \hline
		Discriminator  & \textbf{0.756} & \textbf{0.704} & \textbf{0.623} & \textbf{0.656} & \textbf{0.569} & \textbf{0.469} \\ \hline
		Decoder  & 0.732 & 0.703 & 0.611 & 0.651 & 0.565 & 0.464 \\ \hline
	\end{tabular}
    \end{adjustbox}
    \label{table:discriminator_vs_decoder}
\end{table}

\begin{table}[t]
    \begin{minipage}{0.47\linewidth}
    \centering
    \caption{The decoder architecture for Fashion-MNIST, similarly as the architecture used in~\cite{mukherjee2019clustergan}}
    \begin{adjustbox}{max width=\linewidth}
    \begin{tabular}{c}
		\hline 
		\hline 
		Input $Z=(Z_c,Z_s)$\\ \hline
		Dense, BN 1024 lReLU \\ \hline
		Dense, BN $7\times 7\times 128$ lReLU \\ \hline
		$4\times 4$, stride=2 deconv, BN 64 lReLU \\ \hline
		$4\times 4$, stride=2 deconv, BN 1 tanh \\ \hline
		\hline
	\end{tabular}
    \end{adjustbox}
    \label{table:decoder28}
    \end{minipage}
    \hfill
    \begin{minipage}{0.47\linewidth}
    \centering
    \caption{The decoder architecture for CIFAR-10, similarly as the architecture used in~\cite{gulrajani2017improved} with images converted to grayscale}
    \begin{adjustbox}{max width=\linewidth}
    \begin{tabular}{c}
		\hline 
		\hline 
		Input $Z=(Z_c,Z_s)$\\ \hline
		Dense, $4\times 4\times 512$\\ \hline
		ResBlock up 512 \\ \hline
		ResBlock up 256 \\ \hline
		ResBlock up 128 \\ \hline
		BN, ReLU, $3\times 3$ conv, 1 tanh \\ \hline
		\hline
	\end{tabular}
    \end{adjustbox}
    \label{table:decoder32}
    \end{minipage}
\end{table}

\begin{table}[!th]
    \begin{minipage}{0.47\linewidth}
    \centering
    \caption{The impact of different preprocessing for Fashion-MNIST}
    \begin{adjustbox}{max width=\linewidth}
    \begin{tabular}{|c|ccc|}
		\hline
		Preprocessing & ACC & NMI & ARI  \\ \hline
		None  & 0.756 & 0.704 & 0.623 \\ \hline
		Sobel filtering & \textbf{0.758} & \textbf{0.706} & \textbf{0.625} \\ \hline
	\end{tabular}
    \end{adjustbox}
    \label{table:preprocessing_fashionmnist}
    \end{minipage}
    \hfill
    \begin{minipage}{0.47\linewidth}
    \centering
    \caption{The impact of different preprocessing for CIFAR-10}
    \begin{adjustbox}{max width=\linewidth}
    \begin{tabular}{|c|ccc|ccc|}
		\hline
		Preprocessing & ACC & NMI & ARI  \\ \hline
		None & 0.635 & 0.544 & 0.448 \\ \hline
		Grayscale  & \textbf{0.656} & \textbf{0.569} & \textbf{0.469} \\ \hline
		Sobel filtering & 0.652 & 0.564 & 0.464 \\ \hline
	\end{tabular}
    \end{adjustbox}
    \label{table:preprocessing_cifar10}
    \end{minipage}
\end{table}

\section{Impact of the Image Preprocessing}
For preprocessing, we only convert the color images to grayscale, while IIC~\cite{ji2019invariant} further applies Sobel filtering to extract gradient information. Table~\ref{table:preprocessing_fashionmnist} and Table~\ref{table:preprocessing_cifar10} compare the clustering performance with different preprocessing strategies on Fashion-MNIST and CIFAR-10, respectively. When performing Sobel filtering, a convolutional layer with the Sobel kernel is added before the encoder. For Fashion-MNIST, using Sobel filtering achieves slightly better performance. For CIFAR-10, grayscale without Sobel filtering has the best performance, while clustering on the color images yields the worst performance, indicating that the color information may be trivial for clustering on CIFAR-10.

\begin{table}[t]
    \centering
    \caption{Comparison of the clustering accuracy with other state-of-the-art methods on STL-10~(without the unlablled subset, using ResNet-50~\cite{he2016deep} pretrained with ImageNet~\cite{deng2009imagenet}). The best two results are highlighted in \textbf{bold}}
    \begin{adjustbox}{max width=\linewidth}
    \begin{tabular}{|c|c|}
		\hline
		Method & ACC~(\%) \\ \hline
		AE+GMM~\cite{yang2019dgg} & 79.83 \\ \hline
		DEC~\cite{xie2016unsupervised} & 80.64   \\ \hline
		VaDE~\cite{jiang2016variational} & 84.45 \\ \hline
		RIM~\cite{krause2010discriminative} & 92.50 \\\hline
		IMSAT~\cite{hu2017learning} & \textbf{94.10}  \\\hline
		LTVAE~\cite{li2018learning} & 90.00  \\\hline
		DGG~\cite{yang2019dgg} & 90.59 \\  \hline
		DCCS~(Proposed) & \textbf{95.53} \\\hline
	\end{tabular}
    \end{adjustbox}
    \label{table:stl10_pretrained_res}
\end{table}

\section{Results on STL-10 with Pretrained Model}
Several methods use ResNet-50~\cite{he2016deep} pretrained with ImageNet~\cite{deng2009imagenet} to extract features for clustering. For a fair comparison with these methods, we replace the encoder of DCCS with the same network, \ie a pretrained ResNet-50 followed by three fully-connected layers with 500, 500, 2000 units, respectively. Batch normalization and ReLU activation function are applied on each fully-connected layer. The parameters of the ResNet-50 are fixed during optimization the same as in previous studies. We use RGB images as inputs and resize them to 224 $\times$ 224 pixels. The input $X$ of the discriminator in Table~\ref{table:discriminator} is the average pooled vector of the last residual block of ResNet-50. As shown in Table~\ref{table:stl10_pretrained_res}, DCCS outperforms other state-of-the-art methods, \eg 1.43\% accuracy higher than IMSAT~\cite{hu2017learning}. The NMI and ARI metrics of DCCS are 0.9030 and 0.9051, respectively.

\end{document}